\documentclass[letterpaper]{article} 
\usepackage{aaai2026}  
\nocopyright
\usepackage{times}  
\usepackage{helvet}  
\usepackage{courier}  
\usepackage[hyphens]{url}  
\usepackage{graphicx} 
\usepackage{natbib}  
\usepackage{caption} 
\usepackage{amsmath}
\usepackage{algorithm}
\usepackage{algorithmic}
\usepackage{bibentry}
\usepackage{amssymb}    
\usepackage{color}
\usepackage{multirow}
\usepackage{booktabs}
\usepackage{makecell}
\usepackage{newfloat}
\usepackage{listings}
\usepackage[most]{tcolorbox}
\usepackage{listings}
\tcbset{breakable}
\definecolor{ucblue}{HTML}{003262}
\usepackage[most]{tcolorbox}
\DeclareCaptionStyle{ruled}{labelfont=normalfont,labelsep=colon,strut=off} 
\floatstyle{ruled}
\newfloat{listing}{tb}{lst}{}
\floatname{listing}{Listing}
\usepackage{enumitem}
\setlist[itemize]{leftmargin=*,noitemsep,topsep=0pt}
\setlist[enumerate]{leftmargin=*,noitemsep,topsep=0pt}

\usepackage{microtype}

\title{MV‑Debate: Multi-view Agent Debate with Dynamic Reflection Gating for Multimodal Harmful Content Detection in Social Media }

\author{
  Rui Lu\textsuperscript{\rm 1,2},
  Jinhe Bi\textsuperscript{\rm 3},
  Yunpu Ma\textsuperscript{\rm 3,4},
  Feng Xiao\textsuperscript{\rm 5},
  Yuntao Du\textsuperscript{\rm 1}\thanks{{\raggedright Corresponding authors: yuntaodu@sdu.edu.cn,\\ meetyijun@gmail.com.}},
  Yijun Tian\textsuperscript{\rm 6}\footnotemark[\value{footnote}]
}

\affiliations{
    \textsuperscript{\rm 1}Shandong University\\
    \textsuperscript{\rm 2}Ping An Technology\\
    \textsuperscript{\rm 3}Ludwig Maximilian University of Munich\\
    \textsuperscript{\rm 4}Munich Center for Machine Learning\\
    \textsuperscript{\rm 5}Computility Lab, Beijing EB Technology Co.LTD\\
    \textsuperscript{\rm 6}University of Notre Dame\\
    ruilu42@outlook.com, bijinhe@outlook.com, yunpu.ma@ifi.lmu.de,\\
    fengx@ebtech.com, yuntaodu@sdu.edu.cn, meetyijun@gmail.com
}

\begin{document}

\maketitle

\begin{abstract}

Social media has evolved into a complex multimodal environment where text, images, and other signals interact to shape nuanced meanings, often concealing harmful intent. Identifying such intent, whether sarcasm, hate speech, or misinformation, remains challenging due to cross-modal contradictions, rapid cultural shifts, and subtle pragmatic cues. To address these challenges, we propose MV-Debate, a multi-view agent debate framework with dynamic reflection gating for unified multimodal harmful content detection. MV-Debate assembles four complementary debate agents, a surface analyst, a deep reasoner, a modality contrast, and a social contextualist, to analyze content from diverse interpretive perspectives. Through iterative debate and reflection, the agents refine responses under a $\Delta$-gain criterion, ensuring both accuracy and efficiency. Experiments on three benchmark datasets demonstrate that MV-Debate significantly outperforms strong single-model and existing multi-agent debate baselines. This work highlights the promise of multi-agent debate in advancing reliable social intent detection in safety-critical online contexts.

\end{abstract}
\vspace{-3mm}
\section{Introduction}

The rapid growth of social media platforms as multimodal communication channels—integrating images, short videos, emojis, and stylized texts—has significantly increased the complexity and ambiguity of online messages, posing critical challenges for effective multimodal harmful content detection. For example, multimodal ambiguity occurs when a seemingly neutral caption paired with an ironic image or exaggerated visual cues expresses hidden ridicule, undetectable from text alone; similarly, memes or edited videos frequently amplify emotional or persuasive meanings beyond their literal content, complicating harmful intent detection.  Accurately identifying the underlying \emph{social intent}, whether a post \emph{ridicules} (sarcasm), \emph{vilifies} (hate speech), or \emph{misleads} (misinformation), is thus critical not only for content moderation and community safety, but also for opinion mining, public-discourse analysis, and manipulation-campaign detection. The challenge is heightened by the creative ways users blend modalities, often relying on cultural references, irony, or ambiguity to veil harmful messages. Consequently, effective detection requires integrating linguistic cues, visual semantics, and contextual knowledge to uncover the true communicative intent of multi-modal content, ensuring reliable performance in safety‑critical applications across increasingly complex online environments.

Yet such intent remains challenging to discern because cues are often (i) \emph{cross‑modal}: an image can invert or reinforce a caption’s meaning; (ii) \emph{context‑dependent}: memes, slang, and cultural references evolve rapidly; and (iii) \emph{subtle or sparsely distributed}: harmful intents may be concealed subtly within predominantly benign content.
Empirical studies confirm these obstacles: state-of-the-art multimodal classifiers struggle with culturally grounded irony~\cite{smith2024irony}, while single-stream text-only models like BERT variants become brittle when visual evidence contradicts textual sentiment~\cite{zhang2023visiontext}.

\begin{figure*}[t!]
  \vspace{-3mm}
  \centering
\includegraphics[width=0.84\linewidth]{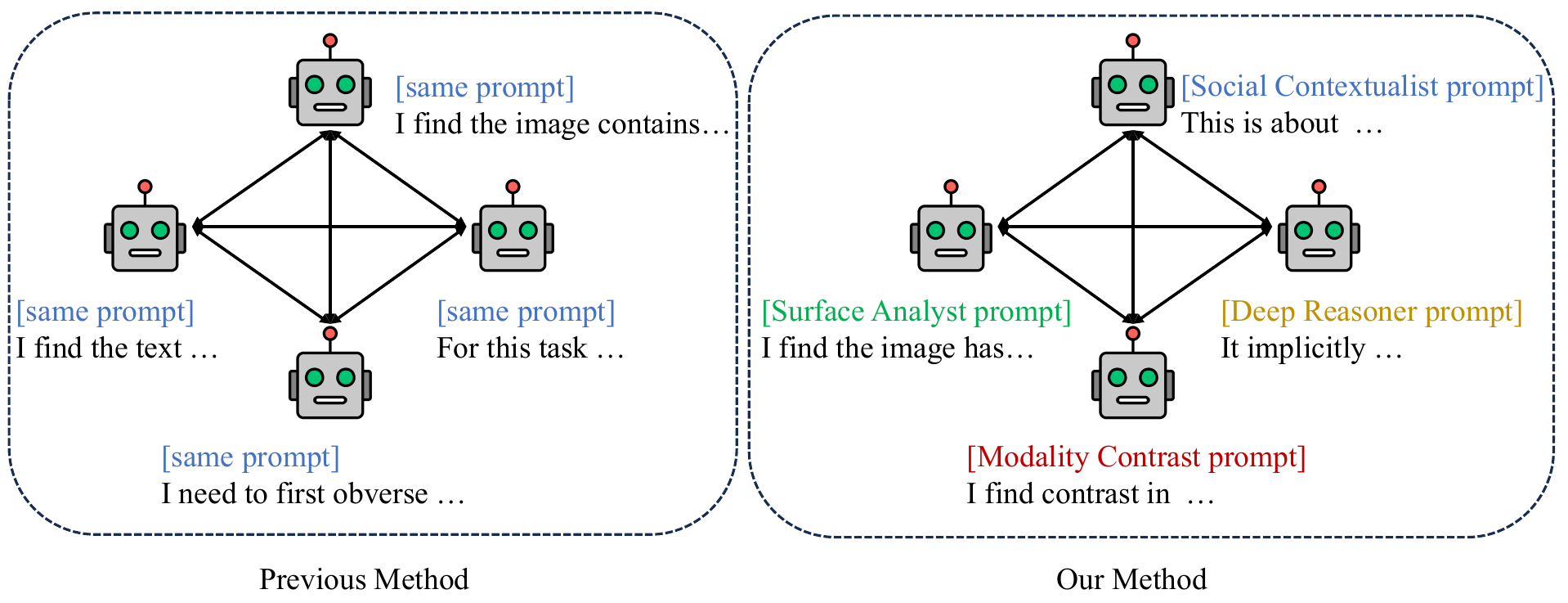}
  \caption{Comparison between existing MAD and our method.}
  \label{fig:motivation}
  \vspace{-6mm}
\end{figure*}

Recently, with the development of LLM agents, multi-agent-based frameworks have achieved remarkable progress in many fields~\cite{guo2024large,chan2023chateval}. Among them, multi-agent debate~\cite{chan2023chateval,Madaan2023SelfRefine} is an effective manner of utilizing the debate among multiple agents to promote the reasoning performance, which can compensate for individual blind spots. Representative methods include opinion-holding~\cite{estornell2024multi} and free-form method~\cite{chan2023chateval}, where the former assigns a predefined opinion (e.g., true for sarcasm task), while the latter performs free-form prediction. Multi-agent debate mechanisms have improved performance on complex reasoning tasks, such as zero-shot stance detection and value-sensitive decision-making, by encouraging agents to expose, defend, and contest alternative interpretations~\cite{Du2023Debate,li2024chainofagents}. Although achieving significant performance, few work has explore multi-agent debate for multimodal harmful content detection.

Current multi-agent debate-based systems typically have the following shortcomings for this scenario: (i) In multi-agent debate settings, existing methods often employ identical LLM prompts across roles, resulting in similar reasoning patterns and repeated errors due to pre-training biases. (ii) The General MAD strategy is designed for general question answering tasks, which often overlooks task-specific design and leads to sub-optimal performance. 
(iii) Existing methods focus on a \emph{single pragmatic category}, which needs multiple different methods for different tasks.

To address these gaps, we propose \textbf{MV-Debate}, a \textbf{M}ulti-\textbf{V}iew agent \textbf{Debate} framework with dynamic reflection gating for unified multimodal harmful content detection.
MV-Debate reformulates three historically siloed tasks—sarcasm~\cite{benchekroun2022mmsd},  hate speech~\cite{kiela2020hateful}, and misinformation detection~\cite{shu2020fakenewsnet}, under a broader \textbf{Harmful Content Detection} objective.
Inspired by the intuition that diverse ways of thinking \cite{liu2025breaking} can promote mutual inspiration, during the debate,  our architecture assembles four different view-based vision–language agents, namely \emph{Surface Analyst}, \emph{Deep Reasoner}, \emph{Modality Contraster}, and \emph{Social Contextualist}.
Each agent adopts a different reasoning perspective and combines insights from the perspective of others.
By comparing these perspectives
and extracting valuable insights, agents refine their answers to reach the correct result.

Specifically, in the first round, given an image and text pair, each agent generates an individual response with a unique given prompting. Then, a judge agent is introduced to score different competing responses, with the better response leading to a higher score.
Next, to better stimulate the potential of each agent, a reflection agent proposes to reflect the top-scoring agents. The new response is adopted only when a measurable \textbf{$\Delta$-gain criterion} (e.g., $\Delta \ge 0.1$ in judge scoring) is met, thus improving the reflection quality. At the second round, each agent reviews and critiques the highest score response selected from the last round, incorporating this feedback to produce its own response. Then, we perform a similar process as the first round. This whole debate process is repeated over several rounds.
MV-Debate promotes diversity by introducing diverse views, leading to more robust solutions.

We conduct extensive experiments on three public benchmarks from three tasks.
The Experimental results on these benchmarks demonstrate that: (i) MV-Debate consistently outperforms strong single-model and existing multi-agent debate baselines across all three intent types;
(ii)  MV-Debate with heterogeneous agents achieves better results than that of homogeneous agents;
(iii) Large model size and more debate rounds often lead to better performance, while they also require more cost and time.

To sum up, the contributions of this work are as follows:
\begin{itemize}
    \item We introduce MV-Debate, a multi-agent debate framework that guides agents to employ diverse reasoning views for multi-modal harmful content detection in social media.
    \item We design four view-specific debate agents with a dynamic reflection gating mechanism to improve the performance.
    \item We empirically validate the effectiveness of the proposed method on multiple multi-modal harmful content benchmarks.
\end{itemize}

\section{Related Work}
\label{sec:related}

\subsection{Harmful Content Detection in Social Media}
Early multimodal studies addressed sarcasm, hate speech, and misinformation as \emph{separate} problems, each with its own dataset and model design. Recent years have shifted toward LMMs and agentic frameworks.

(i) \textbf{Sarcasm}: Tang et al.~\cite{Tang2024GenerativeMLLM} adapt LMMs via visual instruction templates with in-context demonstrations. S³ Agent~\cite{Wang2024S3Agent} integrates multiple LMMs from semantic, superficial, and sentiment views, while Commander-GPT~\cite{Zhang2025CommanderGPT} decomposes sarcasm into six subtasks and aggregates rationales through a central “commander.”

(ii) \textbf{Hate speech}: Van \& Wu~\cite{vanwu2023vlm} show that prompting LLaVA and GPT‑4V with crafted instructions achieves strong zero-shot hateful-meme detection. Yamagishi~\cite{yamagishi2024zeroshot} finds simple prompts outperform complex ones for event detection. Lin~\cite{Lin2024TowardsEH} offers an explainable method by reasoning over conflicting harmless and harmful rationales.

(iii) \textbf{Misinformation}: To tackle scarce and noisy supervision, LVLM4FV~\cite{tahmasebi2024lvlm4fv} combines GPT-ranked evidence retrieval with an InstructBLIP verifier, while SNIFFER~\cite{qi2024sniffer} employs two-stage instruction tuning with entity extraction and image-based web search. LEMMA~\cite{xuan2024lemma} enhances reasoning via multi-query retrieval and distillation. Cekinel et al.~\cite{cekinel2025probing} probe VLM embeddings with a lightweight classifier.

Toward unification of these tasks, MM‑SOC~\cite{Jin2024MMSOC} integrates ten tasks, including sarcasm, hate, and misinformation, revealing LMMs’ fragility in socially nuanced harmful content.

\subsection{Multi‑Agent Debate}
Multi‑agent debate was first shown to improve factual accuracy in open‑domain QA~\cite{Du2023Debate} and later operationalised through open‑source frameworks such as \textsc{AutoGen}~\cite{Wu2023AutoGen}.   Following works studied MAD from different perspectives. Some assign different agents to play different roles \cite{wang2023unleashing}.  There are also other methods improving MAD through embeddings \cite{pham2023let}.
\textsc{ReConcile} arranges a round‑table “conference’’ among LLMs to reach consensus~\cite{Chen2024ReConcile},   while CAF‑I tailors role‑specialised agents to irony detection ~\cite{Liu2025CAFI}. 
These successes confirm that agent heterogeneity and structured interaction benefit hard reasoning tasks.

However, most of them debate with a single thinking which may lead to similar output patterns. Instead, we propose MV-Debate to encourage different agents to
think with distinct reasoning views, which can prompt mutual inspiration. Similarly, \cite{gao2024meta}
dynamically selects the most suitable reasoning method to solve the problem.

\subsection{Large Multimodal Model}

Large Multimodal Models (LMMs) have achieved progress in integrating vision and language, enabling cross-modal understanding and reasoning. A typical LMM comprises three components: a language encoder, a vision encoder, and a cross-modal interaction module~\cite{caffagni2024revolution,liang2022multi11,bi-etal-2025-llava,bi2025cot,bi2025prism}. The cross-modality module bridges the two, allowing effective processing of visual inputs.

Building on this architecture, models such as Qwen2.5-VL~\cite{bai2025qwen2}, InternVL2.5~\cite{chen2024expanding}, and LLaVA series~\cite{liu2023visual,li2024llava} adopt different design choices and training strategies. These advances have significantly improved vision-language alignment, yielding strong performance across multimodal benchmarks~\cite{kil2024mllm,huang2024survey}. Additionally, closed-source models such as GPT-4v, GPT-4o~\cite{hurst2024gpt}, Gemini\cite{comanici2025gemini}, and Claude-Sonnet have demonstrated excellent results in diverse multimodal tasks. Besides, agent-based method~\cite{gao2024clova,fan2024videoagent,wang2025repomaster} also achieved remarkable progress.

\subsection{Reflection in LMMs}
Prompt‑level \emph{self‑reflection} has proven to boost test‑time reasoning.  
\textsc{Self‑Refine}~\cite{Madaan2023SelfRefine} lets a model iteratively critique and rewrite its own answer, improving seven diverse tasks without extra training.  
Renze~\&~Guven~\cite{Renze2024SelfReflection} systematically evaluate eight reflection variants and report significant gains across public question banks.  
Beyond prompting, Bansal~\cite{bensal2025reflect} introduces \textsc{Reflect--Retry--Reward}, a reinforcement‑learning framework that rewards tokens produced during successful reflections. 
Taken together, these studies confirm that reflection is powerful but costly when applied indiscriminately, underscoring the importance of \emph{gated} or selective reflection strategies.

\section{Methodology}
\label{sec:methodology}

This section describes our proposed reflection-gated multi-agent debate framework (\textbf{MV-Debate}) for multimodal harmful content detection in social media.

\subsection{Problem Formulation}
\label{subsec:problem_formulation}

Given a multimodal social-media post composed of text $x^{text}$ and associated visual content $x^\text{img}$, the goal is to predict its underlying \textit{social intent} label $y\in\mathcal{Y}$, where $\mathcal{Y}=\{Yes, No\}$ indicates whether there are sarcasm, hate content, or misinformation.
The objective is to maximize predictive accuracy.

\subsection{System Architecture}
\label{subsec:system_architecture}

Inspired the intuition that diverse reasoning methods could lead to better cooperation results in multi-agent debates. This method could avoid the analogous phenomenon in
existing LMMs, that MAD with a fixed prompting strategy may always produce similar answers to a problem. Thus,
we argue the importance of utilizing different reasoning views in debate to promote diverse thinking
and propose a multi-view-based debate method for multimodal harmful content detection.  We employ a variety of prompting reasoning techniques to produce distinct modes of thought, which do not need training or fine-tuning. We endeavor to design reasoning methods with significant divergence to avoid the issue of similar reasoning processes.

Specially, the MV-Debate system consists of four types of specialized debate agents, alongside three additional control agents. Each debate agent is required to answer the question with the corresponding assigned view, and the control agent aims to score and reflect the reasoning path, and eventually, make a final prediction.

\begin{enumerate}
    \item \textbf{Specialized Debate Agents:}
    \begin{itemize}
        \item \textbf{Surface Analyst agent (SA)}: This agent focuses exclusively on explicit textual and visual cues to detect.
        \item \textbf{Deep Reasoner agent (DR)}: This agent uncovers implicit meanings and hidden intents to detect.
        \item \textbf{Modality Contrast agent (MC)}: This agent assesses alignment or contradictions between textual and visual modalities to detect.
        \item \textbf{Social Contextualist agent (SC)}: This agent leverages external cultural and social-contextual knowledge to detect.
    \end{itemize}
    
    \item \textbf{Judge Agent}: This agent evaluates arguments generated by the debate agents. It assigns scores based on logical coherence, consistency, and plausibility, where a better response would get a higher score.
    
    \item \textbf{Reflection Agent}: This agent generates structured feedback highlighting logical flaws and improvement suggestions.
    
    \item \textbf{Summary Agent}: This agent aggregates the debate history and delivers the final prediction.
\end{enumerate}

\begin{algorithm}[t]
\caption{ Reflection-Gated Multi-View Debate}
\label{alg:mv-debate}
\begin{algorithmic}[1]
    \REQUIRE Input: $(x^{\text{text}}, x^{\text{img}})$, Max rounds $R$, Reflection threshold $\tau$, Top-$k$
    \ENSURE Predicted social intent label: $\hat{y}$
    \STATE Initialize debate history: $H \gets \varnothing$
    \FOR{$t=1$ to $R$}
        \STATE \textbf{Agent responses generation (parallel)}
        \FOR{each agent $i \in \{\text{SA, DR, MC, SC}\}$ in parallel}
            \STATE $\mathbf{r}_{i,t} \gets a.\textsc{Generate}(x^{\text{text}}, x^{\text{img}}, H)$
        \ENDFOR
        \IF{$\textsc{Consensus}(\{\mathbf{r}_{i,t}\})$}
            \RETURN $\textsc{Summary}(H \cup \{\mathbf{r}_{i,t}\})$
        \ENDIF
        \STATE $\mathbf{s}_{i,t} \gets \textsc{Judge}(\{\mathbf{r}_{i,t}\})$ 
        \STATE Update best response: $i^* \gets \arg\max_i s_{i,t}$
        \STATE Append to history: $H \gets H \cup \{(\mathbf{r}_{i^*,t}, s_{i^*,t})\}$
        \STATE Calculate reflection gain:
        \STATE \hspace{1em} $\Delta \gets \textsc{ComputeDelta}(x^{\text{text}}, x^{\text{img}}, k)$
        \IF{$\Delta \geq \tau$} 
            \STATE $\phi_{t} \gets \textsc{Reflect}(H)$
            \STATE Append reflection feedback:
            \STATE  $H \gets H \cup \{\phi_{t}\}$
        \ENDIF
    \ENDFOR
    \RETURN $\textsc{Summary}(H)$
\end{algorithmic}
\end{algorithm}

\subsection{Multi-View Debate}

\subsubsection{Initial Response Generation}
At the first round of debate, given an image-text pair, each specialized debate agent generates its response  $r_{i,1}$, where the subscript “1” means the first round, guided by the corresponding task view prompt $p_i$:
\begin{equation}
    r_{i,1} = M_i(x^{text}, x^{img}|h_i, p_i), i = 1,2,...,4.
\end{equation}
where $h_i$ is the history messages for $i$-th agent and is 
initialized as an empty list. $r_{i,1}$ is output as a structured JSON object comprising a binary decision (`YES` or `NO`) and a brief reasoning. Their role-specific prompts strictly enforce distinct analytical perspectives, ensuring diversity and complementarity in the overall reasoning process.

Consequently, the judge agent collects the solving processes and answers to the questions of all debate agents, and it assigns a score $s_{i,1}$ for each agent's response, with a better response leading to a higher score.

\subsubsection{Top-$k$ $\Delta$-Reflection Gating}
\label{subsec:reflection_gating}

As the initial response from these agents may contain incorrect information, following previous work, we introduce a reflection mechanism to self-improve the response quality.

To reduce computational overhead, we introduce a Top-$k$ $\Delta$-reflection gating strategy. At each round,  the reflection agent would receive all the debate agents' responses and check the reasoning process of each agent. Then, it would point out the reasoning error and provide a revision suggestion. Next, the top $k$ highest responses scored by the judge agent are selected. Then, each selected original debate agent would generate a new response $\hat{r}_{i,1}$ with the query instance, initial response, and revision suggestions. After that, the judge agent would rescore the new response, denoted as  $\hat{s}_{i,1}$. 

Then, we estimate the expected utility of reflection by comparing the scores of the agents with and without reflection feedback. Formally, reflection gain $\Delta_{i,1}$ is calculated as follows:
\begin{equation}
    \Delta_{i,1} = \frac{1}{k}\sum_{i\in \text{Top}_k}\left(\hat{s}_{i,1} - s_{i,1}\right)
\end{equation}
Reflection is only triggered when $\Delta_{i,1}$ surpasses a predefined threshold $\tau$, i.e., $\Delta_{i,1} \geq \tau$. Otherwise, we would use the original response.

In our experiments, we empirically set $k=2$ and $\tau=0.1$ to achieve efficiency improvements. As it could reduce redundant reflection calls by over 60\% compared with reflecting all debate agents, while maintaining or improving accuracy compared to the unconditional reflection baseline.

\subsubsection{History Update} 

After reflection, if the newer response is not adopted, the judge agent would collect highest highest-scoring response, and append it to the history. Otherwise, we will additionally append the reasoning error and revision suggestions $\phi_{1}$ into the history.

\subsubsection{Debate Loop}

Starting from the second round, the best-scoring response from the last round, including both the reasoning processes and the answers, is incorporated into each agent’s history $h_i$. In the following round, each agent leverages these reasoning traces and solutions as additional input, selectively extracting useful information from the diverse perspectives to refine its own answer. This iterative debate process follows the same process as described above, until either the maximum number of rounds $N$ is reached or the agents converge on the same judgment. In our experiments, we set $N=3$.
Finally, at the end of the debate, the summary agent would aggregate the debate history and deliver the final prediction $\hat{y}$. Algorithm~\ref{alg:mv-debate} summarizes the iterative debate and reflection procedure.

\subsection{Discussion}
\label{subsec:discussion}

The proposed reflection-gated multi-view debate framework (MV-Debate) offers three main advantages for multimodal harmful content detection. First, assigning specialized roles to debate agents enforces diverse reasoning perspectives. Unlike single-view prompting, this design combines surface-level, deep semantic, cross-modal, and social-cultural analyses, reducing the risk of missing implicit or context-specific harmful cues.  

Second, the Top-$k$ $\Delta$-reflection gating mechanism enhances reliability while maintaining efficiency. By adaptively triggering reflection only when substantial improvement is expected, the framework avoids redundant computation yet achieves accuracy comparable to or better than unconditional reflection. This is relevant for real-world deployment where scalability and cost-efficiency are critical.  

Third, the iterative debate loop encourages cumulative reasoning. Agents refine their predictions by integrating high-quality responses and structured feedback into their histories, promoting both inter-agent diversity and intra-agent improvement while mitigating repeated errors.

\begin{table*}[t]
\centering 
\vspace{-5mm}
\caption{
The comparison results on three multimodal harmful content detection datasets.
}
\label{res_all}
\resizebox{0.9\textwidth}{!}{ 
\begin{tabular}{c|l|cccccc|cc}
\toprule
   Method    & Model           & \multicolumn{2}{c}{MMSD} & \multicolumn{2}{c}{HatefulMeMe} & \multicolumn{2}{c}{GossipCop} & \multicolumn{2}{c}{Avg} \\
\midrule
   &        & Acc         & F1         & Acc            & F1& Acc           & F1            & Acc        & F1         \\
\midrule
\multirow{11}{*}{Single Model}      & \multicolumn{9}{c}{Closed-source}      \\
\cmidrule(lr){2-10}
          & GPT o4-mini& 77.5    & 78.0   & 70.8       & 63.7       & 72.8      & 66.9      & 73.7   & 69.5   \\
          & GPT 4o     & 78.5    & 75.4   & 75.2       & 71.5       & 73.4      & 75.8      & 75.7   & 74.2   \\
          & Gemini-Flash-2.5        & 73.9    & 80.1   & 77.4       & 67.5       & 76.5      & 72.7      & 75.9   & 73.4   \\
          & Claude-4-Sonnet         & 82.5    & 84.9   & 72.2       & 64.2       & 76.6      & 73.9      & 77.1   & 74.3   \\
         \cmidrule(lr){2-10}
          & \multicolumn{9}{c}{Open-source}        \\
          \cmidrule(lr){2-10}
          & InternVL3-14B           & 74.5    & 78.6   & 68.2       & 64.1       & 72.2      & 68.7      & 71.6   & 70.5   \\
          & Gemma-3-12B& 68.5    & 70.9   & 67.2       & 67.8       & 74.8      & 68.2      & 70.2   & 69.0   \\
          & Qwen2.5-VL-7B-Instruct  & 56.1    & 54.7   & 59.1       & 42.3       & 68.8      & 63.7      & 61.3   & 53.6   \\
          & LLaMA-4-Maverick-17B   & 75.4    & 77.4   & 67.8       & 65.5       & 72.4      & 64.2      & 71.9   & 69.0   \\
\midrule
\multirow{4}{*}{
\makecell{Multi-Agent Debate \\ (Heterogeneous)}} & MAD        & 78.6    & 77.0   & 69.1       & 66.5       & 75.0      & 70.3      & 74.2   & 71.3   \\
          & DMAD       & 81.1    & 81.8   & 72.5       & 69.2       & 77.1      & 72.0      & 76.9   & 74.4   \\
          & ChatEval   & 81.9    & 88.5   & 71.3       & 68.3       & 77.6      & 72.9      & 77.0   & 76.6   \\
          & DebUnc     & 79.6    & 72.1   & 68.6       & 63.2       & 73.8      & 69.6      & 74.0   & 68.3   \\
          \midrule
          & \multicolumn{9}{c}{Open-source}        \\
          \cmidrule(lr){2-10}
\multirow{6}{*}{\makecell{\textbf{Ours} \\ (Homogeneous)}}         & Qwen2.5-VL-7B-instruct  & 65.7    & 62.3   & 61.7       & 61.5       & 71.1      & 65.6      & 66.2   & 63.1   \\
          & InternVL3-14B           & 81.4    & 75.5   & 72.5       & 72.5       & 74.4      & 62.3      & 76.1   & 70.1   \\
          & LLaMA-4-Maverick-17B   & 82.1    & 83.5   & 74.4       & 76.1       & 77.6      & 64.7      & 78.0   & 74.8   \\
          & Gemma-3-12B& 80.4    & 79.1   & 68.3       & 67.5       & 78.1      & 69.8      & 75.6   & 72.1   \\
          \cmidrule(lr){2-10}
          & \multicolumn{9}{c}{Closed-source}      \\
          \cmidrule(lr){2-10}
          & Claude-4-Sonnet         & 90.2    & 86.1   & 80.4       & 70.5       & 78.3      & 69.2      & 82.9   & 75.3   \\
          \midrule
\multirow{2}{*}{\makecell{ \textbf{Ours} \\ (Heterogeneous)}}         & Ours (Open-source)   & 86.1    & 82.5   & 76.0       & 64.5       & 79.4      & 72.3      & 80.5   & 73.1   \\
          & Ours (Closed-source)   & \textbf{92.3}    & \textbf{93.1}   & \textbf{80.8}       & \textbf{70.9}       &  \textbf{81.7} &  70.1&            \textbf{84.9} &   \textbf{78.0} \\
          \bottomrule
\end{tabular}
}
\vspace{-5mm}
\end{table*}

\section{Experiment}
\label{sec:experiment}

\subsection{Setup}

\subsubsection{Datasets}
\label{subsec:datasets}
Following previous work~\cite{lin2024towards,liang2022multi}, in this section, we conduct comprehensive experiments on three widely-used multimodal social context datasets, including the MMSD dataset~\cite{benchekroun2022mmsd} for the sarcasm detection task, the HatefulMeMe dataset~\cite{kiela2020hateful} for the hate speech task, and the GossipCop dataset~\cite{shu2020fakenewsnet} for the misinformation detection task.  As our method and baseline multi-agent debate methods need many tokens for a given instance,  following previous works \cite{Du2023Debate,liu2025breaking}. We do not use the whole dataset and instead randomly select a subset for evaluation.  The number of MMSD, HatefulMeMe, and GossipCop dataset is all 500.

\subsubsection{Baseline Methods} To show the effectiveness of the proposed method, we compare MV-Debate with several types of methods.

The first type of method is the state-of-the-art large multimodal models. For these models, we perform zero-shot prediction with the corresponding task prompt. The selected models includes closed-source models: GPT 4o~\cite{hurst2024gpt}, GPT o4-mini~\cite{openai2025o4mini},  Gemini-Flash-2.5~\cite{comanici2025gemini},  
Claude-4-Sonnet. Besides, we also select some representative open-source models: Qwen2.5-VL~\cite{bai2025qwen2}, InternVL3~\cite{chen2024expanding}, LLaMA-4-Maverick, and Gemma-3~\cite{team2025gemma}.

The second type of method is existing general multi-agent debate methods, including  MAD~\cite{du2023improving}, DMAD~\cite{liu2025breaking},  ChatEval~\cite{chan2023chateval}, and DebUnc~\cite{yoffe2024debunc}. It is noted that these methods are proposed based on LLM and replace the corresponding debate agent with LMMs.

The third type of method is our proposed method and its variants. We implement our methods with both homogeneous and heterogeneous agent scenarios, where the former means that all the debate agent adopts the same LMMs, and the latter adopt different LMMs as the debate agents. For these two scenarios, we test the model on both open-source and closed-source LMMs.

\subsubsection{Implementation Details}

We implement our method based on PyTorch and Huggingface Transformer for the experiments. As for evaluation, we adopt the accuracy and F1 score
as metrics, and all the reported metrics were computed by scikit-learn.  
We utilize closed-source LMMs as our control agents (Judge Agent, Reflection Agent, and Summary Agent)  in MV-Debate, including Claude-4-Sonnet, GPT o4-mini, and GPT 4o, respectively. We set the temperature to 0 and greedy-search to ensure reproducibility. For the Specialized Debate Agents, we use both closed-source and open-source LMMs in our experiments. We use the same model (e.g.,  LLaMA-4-Maverick-17B) for the multi-view specialized agents in the homogeneous experiments. For the heterogeneous settings, we use four unique LMMs as our multi-view specialized agents. We leveraged API interfaces to invoke closed-source LMMs and implemented an asynchronous strategy to execute specialized debate agents in parallel, significantly enhancing runtime efficiency. The experimental hyperparameters in the code fall into three main categories: 
(i) Randomness: We set the random seed to 42 in all experiments. (ii) Debate process control: we set max rounds $N=3$, reflection-gain threshold $\tau=0.1$, and $ k = \lfloor \frac{L}{2} \rfloor$ (where $L$ represents the number of multi-view specialized agents) as the top-$k$ agents selected for the computation of reflection gain. (iii) API scheduling: we set max retries $p = 5 $ times and retry delay $q = 3$  seconds for all the closed-source LMM agents in our experiments.

\subsection{Main Results}

The comparison with baseline methods is shown in Table~\ref{res_all}. Based on the results, we have the following findings.

\textbf{Single-model baselines.} Closed-source models generally outperform open-source counterparts. Claude-4-Sonnet achieves the best overall performance among single models, while GPT 4o and GPT o4-mini trail slightly behind. In contrast, open-source models show a notable performance gap, with Qwen2.5-VL variants performing poorly. These results indicate the difficulty of applying off-the-shelf open-source LMMs to harmful content detection.

\textbf{Existing multi-agent debate baselines.} 
We implement the existing multi-agent debate baseline in an 
heterogeneous
setting with SOTA open-source LMMs (
including Qwen2.5-VL-7B-instruct, InternVL3-14B,  
LLaMA-4-Maverick-17B, and
Gemma-3-12B). The results show that 
existing multi-agent debate frameworks (e.g., DMAD, ChatEval) demonstrate clear advantages over single models. ChatEval, for example, achieves 76.6\% of F1 score, confirming that multi-agent collaboration with a debate manner improves robustness.

\textbf{Our homogeneous MV-Debate.} 
In a homogeneous scenario, we implement MV-Debate with four kinds of open-source LMMs. The results show that our homogeneous framework consistently improves over its base models. For instance, the accuracy of Gemma-3-12B increases from 70.2 (single) to 75.6 with MV-Debate, while InternVL3-14B improves from 71.6 to 76.1. LLaMA-4-Maverick-17B achieves the best open-source results with an accuracy of 78.0. These gains validate the effectiveness of enforcing multi-view reasoning and reflection even without heterogeneous agents.  

\textbf{Our heterogeneous MV-Debate.} 
We implement MV-Debate in a heterogeneous scenario with both
open- and closed-source LMMs, and our heterogeneous framework achieves the highest performance. Besides, the open-source variant yields an accuracy of 80.5, which outperforms existing multi-agent debate methods.
The closed-source variant reaches an accuracy of 84.9, surpassing all baselines. This demonstrates that reflection-gated multi-view debate establishes new state-of-the-art results.  Besides, the heterogeneous agent achieves better results than a homogenous agent under open-source models, which implies that different models could also lead to diverse thinking.

In summary, MV-Debate effectively integrates diverse reasoning views with adaptive reflection, achieving superior accuracy and efficiency. The results highlight its promise as a scalable and reliable framework for multimodal harmful content detection.

\subsection{Insightful Analysis}

\subsubsection{Ablation study of debate agents}

We conduct an ablation study to evaluate the contribution of each specialized debate agent in the homogeneous scenario, as the heterogeneous scenario would couple the effect of different LMMs. We evaluate four agents: Surface Analyst (SA), Deep Reasoner (DR), Modality Contraster (MC), and Social Contextualist (SC). The results are shown in Table~\ref{ab_1}.  

For LLaMA-4-Maverick-17B, removing any agent leads to a performance drop compared with the full MV-Debate. The most significant decline occurs when excluding the Modality Contrast. This confirms the critical role of assessing consistency and contradictions between modalities. Removing the deep reasoner also causes a large drop, highlighting the importance of capturing implicit meanings and hidden harmful intents that are often overlooked by surface-level cues. Excluding the Social Contextualist results in a moderate decline, suggesting that external sociocultural knowledge is essential to interpret nuanced harmful signals. In contrast, removing the Surface Analyst causes only a minor drop, as explicit cues may be partly covered by other agents.  Besides,  a similar trend is observed for Claude-4-Sonnet. 
These consistent patterns across both open- and closed-source models validate the necessity of combining diverse reasoning views.

\begin{table}[t]
\centering 
\caption{
Ablation study about the debate agent.
}
\label{ab_1}
\resizebox{0.46\textwidth}{!}{
\begin{tabular}{c|cc|cc}
\toprule
    Settings   & \multicolumn{2}{c}{LLaMA-4-Maverick-17B} & \multicolumn{2}{c}{Claude-4-Sonnet} \\
\midrule
& Acc	& F1 &	Acc	& F1 \\
\cmidrule{2-5}
\textbf{Ours}               & 82.1              & 83.5              & 90.2           & 86.1           \\
\textbf{w/o SA}    & 80.4              & 87.2              & 87.6           & 85.1           \\
\textbf{w/o DR}    & 78.4              & 69.8              & 86.3           & 84.4           \\
\textbf{w/o MC}    & 75.7              & 72.5              & 85.7           & 83.5           \\
\textbf{w/o SC}    & 77.5              & 74.5              & 86.1           & 82.5           \\
\textbf{Zero-shot} & 75.4              & 77.4              & 82.5           & 84.9          \\
\bottomrule
\end{tabular}
}
 \vspace{-2mm}
\end{table}

\subsubsection{Ablation Study of reflection mechanism}

Table~\ref{ab_2} demonstrates the impact of incorporating reflection in our method in both homogeneous and heterogeneous scenarios.
Across all models, reflection consistently improves both accuracy and F1. For example, for Claude-4-Sonet, the accuracy increases from 85.1 to 90.2 (+5.3), highlighting its effectiveness in correcting reasoning errors and enhancing the detection of implicit harmful cues. 
In summary, reflection plays a pivotal role in maximizing the effectiveness of multi-agent debate. By selectively guiding agents to revise their reasoning, it stabilizes outputs and improves consistency, making the framework more reliable and scalable for real-world harmful content detection tasks.

\begin{table}[h]
\centering 
\caption{
Ablation study about reflection.
"homo" and "hete" mean homogeneous and heterogeneous, respectively.}
\label{ab_2}
\resizebox{0.48\textwidth}{!}{ 
\begin{tabular}{c|l|cc|cc}
\toprule
  \multicolumn{2}{c}{} & \multicolumn{2}{c}{w/o Reflection} & \multicolumn{2}{c}{with Reflection} \\
  \midrule
\multicolumn{1}{l|}{} & \textbf{}& Acc              & F1              & Acc              & F1               \\
\cmidrule{2-6}
\multirow{2}{*}{\makecell{ \textbf{Ours} \\(homo)}} & LLaMA-4-maverick-17B    & 80.4           & 78.2          & 82.1           & 83.5           \\
& Claude-4-Sonnet & 85.1           & 82.3          & 90.2           & 86.1           \\
\cmidrule{2-6}
\multirow{2}{*}{\makecell{ \textbf{Ours} \\(hete)}} &Ours(Open-source) & 84.3           & 79.5          & 86.1           & 82.5           \\
& Ours(Closed-source) & \textbf{88.2}  & \textbf{87.5}  & \textbf{92.3}  & \textbf{93.1}  \\
\bottomrule
\end{tabular}
}
 \vspace{-2mm}
\end{table}

\subsubsection{Ablation Study of the best history}

During the debate, we would select the best-scoring response of the debate agents instead of that of all agents. To show the effectiveness, we compare these two settings in both homogeneous and heterogeneous scenarios. The results of the ablation study of the best history are shown in Table \ref{ab_3}. Our method with LLaMA-4-Maverick-17B shows notable sensitivity to data quality, with its accuracy improving from 70.1 ("All History") to 82.1 ("Best History"), suggesting it benefits substantially from filtered or higher-quality data.  A consistent trend across all models is the superior performance in the "Best History" setting compared to "All History," highlighting the importance of data quality and curation. This effect is most pronounced for LLaMA-4-Maverick, suggesting it is particularly vulnerable to noisy or suboptimal data. 


\begin{table}[h]
\centering 
\caption{
Ablation study about history. "homo" and "hete" mean homogeneous and heterogeneous, respectively.
}
\label{ab_3}
\resizebox{0.48\textwidth}{!}{ 
\begin{tabular}{c|l|cc|cc}
\toprule
\multicolumn{2}{l}{\textbf{}} & \multicolumn{2}{c}{All History} & \multicolumn{2}{c}{Best History} \\
\midrule
\multicolumn{1}{l|}{} & \textbf{}& Acc              & F1              & Acc              & F1               \\
\cmidrule{2-6}
\multirow{2}{*}{\makecell{\textbf{Ours} \\ (homo)}} & LLaMA-4-Maverick(17B)         & 70.1          & 62.8         & 82.1          & 83.5         \\
& Claude-4-Sonnet      & 80.4          & 78.6         & 90.2          & 86.1         \\
\cmidrule{2-6}
\multirow{2}{*}{\makecell{\textbf{Ours} \\(hete)}} & Ours (Open-source)   & 72.0& 63.5& 86.1   & 82.5       \\
& Ours (Closed-source)  & \textbf{82.2} &  \textbf{80.1}& \textbf{92.3}   & \textbf{93.1}    \\
\bottomrule
\end{tabular}
}
\vspace{-5mm}
\end{table}

\subsubsection{Ablation about debate round}

\begin{figure}[h]
\vspace{-3mm}
  \centering
\includegraphics[width=0.8\linewidth]{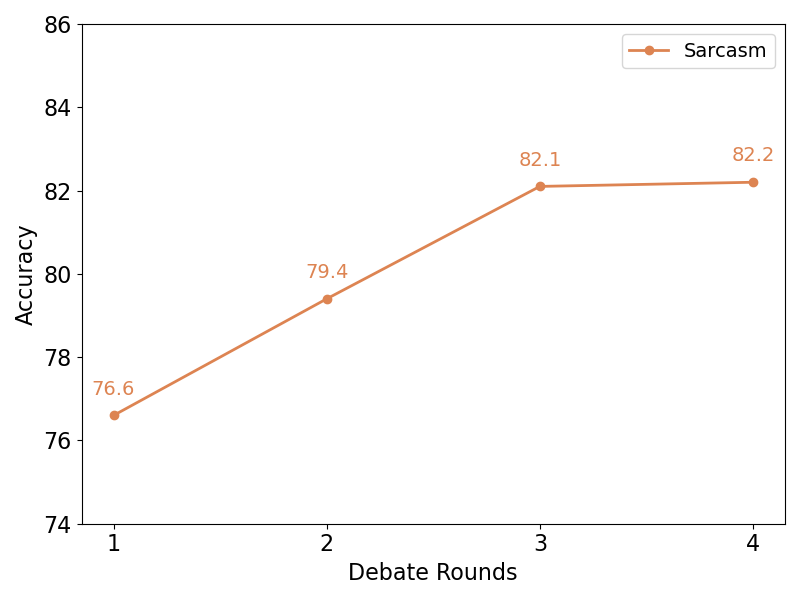}
  \caption{Ablation about debate round.}
  \label{fig:round}
  \vspace{-2mm}
\end{figure}

The effect of varying the number of debate rounds from 1 to 4 in a homogeneous scenario based on LLaMA-4-maverick-17B is shown in Figure~\ref{fig:round}. 
We observe that increasing the debate rounds generally enhances performance. Specifically, accuracy rises from 76.6 at one round to 82.1 at four rounds. Notably, the transition from one to three rounds yields the most substantial gains. The fourth round brings marginal improvements, suggesting that iterative debate allows agents to progressively refine their reasoning by integrating complementary perspectives, though performance gains tend to saturate after three rounds. These results highlight the effectiveness of multi-round debate in enhancing both robustness and precision, while indicating a practical trade-off between performance gains and additional computational overhead. Thus, considering both efficiency and effectiveness, set the debate round to be 3 in our experiments.

\subsubsection{Ablation about model size}

\begin{figure}[h]
  \vspace{-2mm}
  \centering
\includegraphics[width=0.9\linewidth]{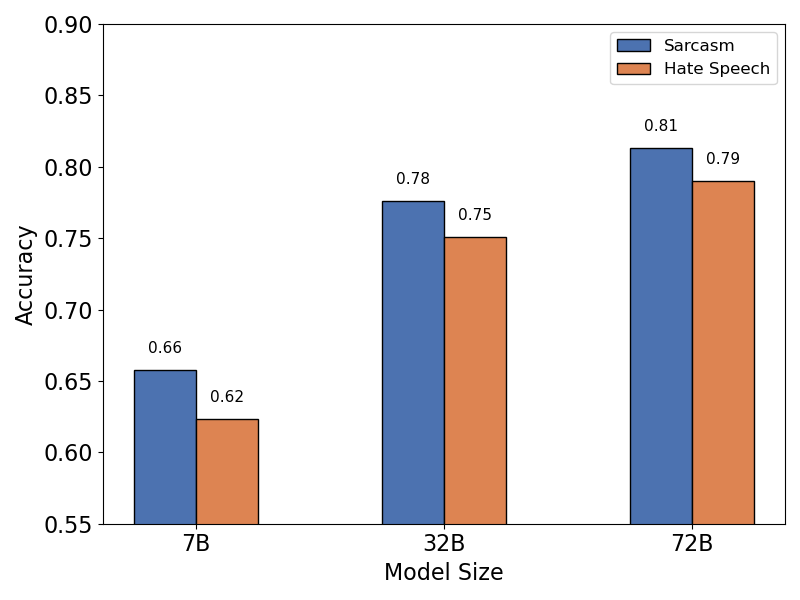}
  \caption{Ablation about model size on Qwen2.5-VL.}
  \label{fig:size}
  \vspace{-2mm}
\end{figure}

The results of the impact of model size on sarcasm and hate detection in a homogeneous scenario on Qwen2.5-VL series models are shown in  Table~\ref{fig:size}. As the parameter scale increases from 7B to 72B, the accuracy consistently improves across tasks. For sarcasm detection, accuracy rises from 66\% to 81\%, indicating a substantial gain of over 15 percentage points. Similarly, for hate detection, accuracy increases from 62\% to 79\%. These results suggest that larger models possess a stronger capacity for capturing subtle multimodal cues and complex pragmatic signals, leading to more accurate social intent classification.

\section{Conclusion}

In this work, we introduced MV-Debate, a novel multi-view debate framework for multimodal harmful content detection on social media. By orchestrating four view-specific agents with complementary reasoning strategies and a dynamic reflection gating mechanism, MV-Debate effectively integrates cross-modal evidence and contextual cues to identify complex social intents such as sarcasm, hate speech, and misinformation. Extensive experiments across multiple benchmarks confirm its superior accuracy, efficiency, and interpretability compared with strong baselines. Beyond performance gains, MV-Debate also generates transparent debate transcripts, supporting model debugging, auditing, and user trust. Looking forward, our framework provides a foundation for extending multi-agent debate approaches to broader safety-critical multimodal reasoning tasks.

As for the limitation, the framework’s performance depends on the underlying LMMs, which may inherit biases or struggle with culturally nuanced content such as sarcasm. The current design also fixes the number of reasoning views, which may not always balance accuracy and efficiency.

\newpage

\bibliography{aaai2026}

\begin{thebibliography}{53}
\providecommand{\natexlab}[1]{#1}

\bibitem[{Bai et~al.(2025)Bai, Chen, Liu, Wang, Ge, Song, Dang, Wang, Wang, Tang et~al.}]{bai2025qwen2}
Bai, S.; Chen, K.; Liu, X.; Wang, J.; Ge, W.; Song, S.; Dang, K.; Wang, P.; Wang, S.; Tang, J.; et~al. 2025.
\newblock Qwen2. 5-vl technical report.
\newblock \emph{arXiv preprint arXiv:2502.13923}.

\bibitem[{Balauca et~al.(2025)Balauca, Garai, Balauca, Shetty, Agrawal, Shah, Fu, Wang, Toutanova, Paudel, and Gool}]{liang2022multi11}
Balauca, A.-A.; Garai, S.; Balauca, S.; Shetty, R.~U.; Agrawal, N.; Shah, D.~S.; Fu, Y.; Wang, X.; Toutanova, K.; Paudel, D.~P.; and Gool, L.~V. 2025.
\newblock Understanding the World's Museums through Vision-Language Reasoning.
\newblock In \emph{ICCV}.

\bibitem[{Benchekroun et~al.(2022)Benchekroun, Istrate, Zalc, and Lenne}]{benchekroun2022mmsd}
Benchekroun, M.; Istrate, D.; Zalc, V.; and Lenne, D. 2022.
\newblock Mmsd: A Multi-modal Dataset for Real-time, Continuous Stress Detection from Physiological Signals.
\newblock In \emph{HEALTHINF}, 240--248.

\bibitem[{Bensal et~al.(2025)Bensal, Jamil, Bryant, Russak, Kamble, Mozolevskyi, Ali, and AlShikh}]{bensal2025reflect}
Bensal, S.; Jamil, U.; Bryant, C.; Russak, M.; Kamble, K.; Mozolevskyi, D.; Ali, M.; and AlShikh, W. 2025.
\newblock Reflect, Retry, Reward: Self-Improving LLMs via Reinforcement Learning.
\newblock \emph{arXiv preprint arXiv:2505.24726}.

\bibitem[{Bi et~al.(2025)Bi, Wang, Chen, Xiao, Hecker, Tresp, and Ma}]{bi-etal-2025-llava}
Bi, J.; Wang, Y.; Chen, H.; Xiao, X.; Hecker, A.; Tresp, V.; and Ma, Y. 2025.
\newblock LLaVA Steering: Visual Instruction Tuning with 500x Fewer Parameters through Modality Linear Representation-Steering.
\newblock In \emph{Proceedings of the 63rd Annual Meeting of the Association for Computational Linguistics}, 15230--15250.

\bibitem[{Caffagni et~al.(2024)Caffagni, Cocchi, Barsellotti, Moratelli, Sarto, Baraldi, Cornia, and Cucchiara}]{caffagni2024revolution}
Caffagni, D.; Cocchi, F.; Barsellotti, L.; Moratelli, N.; Sarto, S.; Baraldi, L.; Cornia, M.; and Cucchiara, R. 2024.
\newblock The revolution of multimodal large language models: a survey.
\newblock \emph{ACL}.

\bibitem[{Cekinel, Karagoz, and oltekin(2025)}]{cekinel2025probing}
Cekinel, R.~F.; Karagoz, P.; and oltekin. 2025.
\newblock Multimodal Fact-Checking with Vision Language Models: A Probing Classifier Based Solution with Embedding Strategies.
\newblock In \emph{Proceedings of COLING}.

\bibitem[{Chan et~al.(2023)Chan, Chen, Su, Yu, Xue, Zhang, Fu, and Liu}]{chan2023chateval}
Chan, C.-M.; Chen, W.; Su, Y.; Yu, J.; Xue, W.; Zhang, S.; Fu, J.; and Liu, Z. 2023.
\newblock Chateval: Towards better llm-based evaluators through multi-agent debate.
\newblock \emph{arXiv preprint arXiv:2308.07201}.

\bibitem[{Chen, Saha, and Bansal(2024)}]{Chen2024ReConcile}
Chen, J.; Saha, S.; and Bansal, M. 2024.
\newblock ReConcile: Round‑Table Conference Improves Reasoning via Consensus among Diverse {LLMs}.
\newblock In \emph{Proc. ACL 2024}, 7066--7085.

\bibitem[{Chen et~al.(2024)Chen, Wang, Cao, Liu, Gao, Cui, Zhu, Ye, Tian, Liu et~al.}]{chen2024expanding}
Chen, Z.; Wang, W.; Cao, Y.; Liu, Y.; Gao, Z.; Cui, E.; Zhu, J.; Ye, S.; Tian, H.; Liu, Z.; et~al. 2024.
\newblock Expanding performance boundaries of open-source multimodal models with model, data, and test-time scaling.
\newblock \emph{arXiv preprint arXiv:2412.05271}.

\bibitem[{Comanici et~al.(2025)Comanici, Bieber, Schaekermann, Pasupat, Sachdeva, Dhillon, Blistein, Ram, Zhang, Rosen et~al.}]{comanici2025gemini}
Comanici, G.; Bieber, E.; Schaekermann, M.; Pasupat, I.; Sachdeva, N.; Dhillon, I.; Blistein, M.; Ram, O.; Zhang, D.; Rosen, E.; et~al. 2025.
\newblock Gemini 2.5: Pushing the frontier with advanced reasoning, multimodality, long context, and next generation agentic capabilities.
\newblock \emph{arXiv preprint arXiv:2507.06261}.

\bibitem[{Du et~al.(2023{\natexlab{a}})Du, Li, Torralba, Tenenbaum, and Igor Mordatch}]{Du2023Debate}
Du, Y.; Li, S.; Torralba, A.; Tenenbaum, J.; and Igor Mordatch. 2023{\natexlab{a}}.
\newblock Improving Factuality and Reasoning in Language Models through Multiagent Debate.
\newblock \emph{arXiv preprint}, arXiv:2305.14325.

\bibitem[{Du et~al.(2023{\natexlab{b}})Du, Li, Torralba, Tenenbaum, and Mordatch}]{du2023improving}
Du, Y.; Li, S.; Torralba, A.; Tenenbaum, J.~B.; and Mordatch, I. 2023{\natexlab{b}}.
\newblock Improving factuality and reasoning in language models through multiagent debate.
\newblock In \emph{Forty-first International Conference on Machine Learning}.

\bibitem[{Estornell and Liu(2024)}]{estornell2024multi}
Estornell, A.; and Liu, Y. 2024.
\newblock Multi-LLM debate: Framework, principals, and interventions.
\newblock \emph{Advances in Neural Information Processing Systems}, 37: 28938--28964.

\bibitem[{Fan et~al.(2024)Fan, Ma, Wu, Du, Li, Gao, and Li}]{fan2024videoagent}
Fan, Y.; Ma, X.; Wu, R.; Du, Y.; Li, J.; Gao, Z.; and Li, Q. 2024.
\newblock Videoagent: A memory-augmented multimodal agent for video understanding.
\newblock In \emph{European Conference on Computer Vision}, 75--92. Springer.

\bibitem[{Gao et~al.(2024{\natexlab{a}})Gao, Xie, Mao, Wu, Xia, Mi, and Wei}]{gao2024meta}
Gao, P.; Xie, A.; Mao, S.; Wu, W.; Xia, Y.; Mi, H.; and Wei, F. 2024{\natexlab{a}}.
\newblock Meta reasoning for large language models.
\newblock \emph{arXiv preprint arXiv:2406.11698}.

\bibitem[{Gao et~al.(2024{\natexlab{b}})Gao, Du, Zhang, Ma, Han, Zhu, and Li}]{gao2024clova}
Gao, Z.; Du, Y.; Zhang, X.; Ma, X.; Han, W.; Zhu, S.-C.; and Li, Q. 2024{\natexlab{b}}.
\newblock Clova: A closed-loop visual assistant with tool usage and update.
\newblock In \emph{Proceedings of the IEEE/CVF conference on computer vision and pattern recognition}, 13258--13268.

\bibitem[{Guo et~al.(2024)Guo, Chen, Wang, Chang, Pei, Chawla, Wiest, and Zhang}]{guo2024large}
Guo, T.; Chen, X.; Wang, Y.; Chang, R.; Pei, S.; Chawla, N.~V.; Wiest, O.; and Zhang, X. 2024.
\newblock Large language model based multi-agents: A survey of progress and challenges.
\newblock \emph{arXiv preprint arXiv:2402.01680}.

\bibitem[{Hee, Chong, and Lee(2023)}]{zhang2023visiontext}
Hee, M.; Chong, W.; and Lee, R. 2023.
\newblock Decoding the Underlying Meaning of Multimodal Hateful Memes.
\newblock In \emph{Proceedings of the 32nd International Joint Conference on Artificial Intelligence (IJCAI 2023)}, 5995--6002.

\bibitem[{Huang and Zhang(2024)}]{huang2024survey}
Huang, J.; and Zhang, J. 2024.
\newblock A survey on evaluation of multimodal large language models.
\newblock \emph{arXiv preprint arXiv:2408.15769}.

\bibitem[{Hurst et~al.(2024)Hurst, Lerer, Goucher, Perelman, Ramesh, Clark, Ostrow, Welihinda, Hayes, Radford et~al.}]{hurst2024gpt}
Hurst, A.; Lerer, A.; Goucher, A.~P.; Perelman, A.; Ramesh, A.; Clark, A.; Ostrow, A.; Welihinda, A.; Hayes, A.; Radford, A.; et~al. 2024.
\newblock Gpt-4o system card.
\newblock \emph{arXiv preprint arXiv:2410.21276}.

\bibitem[{Jin et~al.(2024)Jin, Choi, Verma, Wang, and Kumar}]{Jin2024MMSOC}
Jin, Y.; Choi, M.; Verma, G.; Wang, J.; and Kumar, S. 2024.
\newblock {MM-SOC}: Benchmarking Multimodal Large Language Models in Social Media Platforms.
\newblock In \emph{Findings of ACL 2024}, 6192--6210.

\bibitem[{Jinhe et~al.(2025{\natexlab{a}})Jinhe, Wang, Yan, Xiao, Hecker, Tresp, and Ma}]{bi2025prism}
Jinhe, B.; Wang, Y.; Yan, D.; Xiao, X.; Hecker, A.; Tresp, V.; and Ma, Y. 2025{\natexlab{a}}.
\newblock Prism: Self-pruning intrinsic selection method for training-free multimodal data selection.
\newblock \emph{arXiv preprint arXiv:2502.12119}.

\bibitem[{Jinhe et~al.(2025{\natexlab{b}})Jinhe, Yan, Wang, Huang, Chen, Wan, Ye, Xiao, Schuetze, Tresp et~al.}]{bi2025cot}
Jinhe, B.; Yan, D.; Wang, Y.; Huang, W.; Chen, H.; Wan, G.; Ye, M.; Xiao, X.; Schuetze, H.; Tresp, V.; et~al. 2025{\natexlab{b}}.
\newblock CoT-Kinetics: A Theoretical Modeling Assessing LRM Reasoning Process.
\newblock \emph{arXiv preprint arXiv:2505.13408}.

\bibitem[{Kiela et~al.(2020)Kiela, Firooz, Mohan, Goswami, Singh, Ringshia, and Testuggine}]{kiela2020hateful}
Kiela, D.; Firooz, H.; Mohan, A.; Goswami, V.; Singh, A.; Ringshia, P.; and Testuggine, D. 2020.
\newblock The hateful memes challenge: Detecting hate speech in multimodal memes.
\newblock \emph{Advances in neural information processing systems}, 33: 2611--2624.

\bibitem[{Kil et~al.(2024)Kil, Mai, Lee, Chowdhury, Wang, Cheng, Wang, Liu, and Chao}]{kil2024mllm}
Kil, J.; Mai, Z.; Lee, J.; Chowdhury, A.; Wang, Z.; Cheng, K.; Wang, L.; Liu, Y.; and Chao, W.-L.~H. 2024.
\newblock Mllm-compbench: A comparative reasoning benchmark for multimodal llms.
\newblock \emph{Advances in Neural Information Processing Systems}, 37: 28798--28827.

\bibitem[{Li et~al.(2025)Li, Zhang, Guo, Zhang, Li, Zhang, Zhang, Zhang, Li, Liu et~al.}]{li2024llava}
Li, B.; Zhang, Y.; Guo, D.; Zhang, R.; Li, F.; Zhang, H.; Zhang, K.; Zhang, P.; Li, Y.; Liu, Z.; et~al. 2025.
\newblock Llava-onevision: Easy visual task transfer.
\newblock \emph{Transactions on Machine Learning Research}.

\bibitem[{Liang et~al.(2022)Liang, Lou, Li, Yang, Gui, He, Pei, and Xu}]{liang2022multi}
Liang, B.; Lou, C.; Li, X.; Yang, M.; Gui, L.; He, Y.; Pei, W.; and Xu, R. 2022.
\newblock Multi-modal sarcasm detection via cross-modal graph convolutional network.
\newblock In \emph{Proceedings of the 60th Annual Meeting of the Association for Computational Linguistics (Volume 1: Long Papers)}, volume~1, 1767--1777. Association for Computational Linguistics.

\bibitem[{Lin et~al.(2024{\natexlab{a}})Lin, Luo, Gao, Ma, Wang, and Yang}]{Lin2024TowardsEH}
Lin, H.; Luo, Z.; Gao, W.; Ma, J.; Wang, B.; and Yang, R. 2024{\natexlab{a}}.
\newblock Towards Explainable Harmful Meme Detection through Multimodal Debate between Large Language Models.
\newblock \emph{Proceedings of the ACM Web Conference 2024}.

\bibitem[{Lin et~al.(2024{\natexlab{b}})Lin, Luo, Gao, Ma, Wang, and Yang}]{lin2024towards}
Lin, H.; Luo, Z.; Gao, W.; Ma, J.; Wang, B.; and Yang, R. 2024{\natexlab{b}}.
\newblock Towards explainable harmful meme detection through multimodal debate between large language models.
\newblock In \emph{Proceedings of the ACM Web Conference 2024}, 2359--2370.

\bibitem[{Liu et~al.(2023)Liu, Li, Wu, and Lee}]{liu2023visual}
Liu, H.; Li, C.; Wu, Q.; and Lee, Y.~J. 2023.
\newblock Visual instruction tuning.
\newblock \emph{Advances in neural information processing systems}, 36: 34892--34916.

\bibitem[{Liu et~al.(2025)Liu, Cao, Li, He, and Tan}]{liu2025breaking}
Liu, Y.; Cao, J.; Li, Z.; He, R.; and Tan, T. 2025.
\newblock Breaking mental set to improve reasoning through diverse multi-agent debate.
\newblock In \emph{The Thirteenth International Conference on Learning Representations}.

\bibitem[{Liu, Zhou, and Hu(2025)}]{Liu2025CAFI}
Liu, Z.; Zhou, Z.; and Hu, M. 2025.
\newblock {CAF-I}: A Collaborative Multi‑Agent Framework for Enhanced Irony Detection with Large Language Models.
\newblock \emph{arXiv preprint}, arXiv:2506.08430.

\bibitem[{Lu et~al.(2025)Lu, Dong, Guo, Zhang, Lu, Wang, and Zhang}]{smith2024irony}
Lu, M.; Dong, Z.; Guo, Z.; Zhang, X.; Lu, X.; Wang, T.; and Zhang, L. 2025.
\newblock A Multi-Modal Sarcasm Detection Model Based on Cue Learning.
\newblock \emph{Scientific Reports}, 15(10261).
\newblock Open‐access article.

\bibitem[{Madaan et~al.(2023)Madaan, Tandon, Gupta, Hallinan, Gao, and {\em et al.}}]{Madaan2023SelfRefine}
Madaan, A.; Tandon, N.; Gupta, P.; Hallinan, S.; Gao, L.; and {\em et al.} 2023.
\newblock Self‑Refine: Iterative Refinement with Self‑Feedback.
\newblock \emph{arXiv preprint}, arXiv:2303.17651.

\bibitem[{{OpenAI}(2025)}]{openai2025o4mini}
{OpenAI}. 2025.
\newblock Introducing OpenAI o3 and o4-mini.
\newblock \url{https://openai.com/index/introducing-o3-and-o4-mini}.
\newblock Accessed 2 Aug 2025.

\bibitem[{Pham et~al.(2023)Pham, Liu, Yang, Chen, Liu, Yuan, Plummer, Wang, and Yang}]{pham2023let}
Pham, C.; Liu, B.; Yang, Y.; Chen, Z.; Liu, T.; Yuan, J.; Plummer, B.~A.; Wang, Z.; and Yang, H. 2023.
\newblock Let models speak ciphers: Multiagent debate through embeddings.
\newblock \emph{arXiv preprint arXiv:2310.06272}.

\bibitem[{Qi et~al.(2024)Qi, Yan, Hsu, and Lee}]{qi2024sniffer}
Qi, P.; Yan, Z.; Hsu, W.; and Lee, M.~L. 2024.
\newblock SNIFFER: Multimodal Large Language Model for Explainable Out-of-Context Misinformation Detection.
\newblock In \emph{Proceedings of CVPR}.

\bibitem[{Renze and Guven(2024)}]{Renze2024SelfReflection}
Renze, M.; and Guven, E. 2024.
\newblock Self‑Reflection in {LLM} Agents: Effects on Problem‑Solving Performance.
\newblock \emph{Proceedings of FLLM 2024}.

\bibitem[{Shu et~al.(2020)Shu, Mahudeswaran, Wang, Lee, and Liu}]{shu2020fakenewsnet}
Shu, K.; Mahudeswaran, D.; Wang, S.; Lee, D.; and Liu, H. 2020.
\newblock Fakenewsnet: A data repository with news content, social context, and spatiotemporal information for studying fake news on social media.
\newblock \emph{Big data}, 8(3): 171--188.

\bibitem[{Tahmasebi, M{\"u}ller-Budack, and Ewerth(2024)}]{tahmasebi2024lvlm4fv}
Tahmasebi, S.; M{\"u}ller-Budack, E.; and Ewerth, R. 2024.
\newblock Multimodal Misinformation Detection using Large Vision-Language Models.
\newblock In \emph{Proceedings of CIKM}.

\bibitem[{Tang et~al.(2024)Tang, Lin, Yan, and Li}]{Tang2024GenerativeMLLM}
Tang, B.; Lin, B.; Yan, H.; and Li, S. 2024.
\newblock Leveraging Generative Large Language Models with Visual Instruction and Demonstration Retrieval for Multimodal Sarcasm Detection.
\newblock In \emph{Proceedings of the 2024 Conference of the North American Chapter of the Association for Computational Linguistics: Human Language Technologies}, 1732--1742. Mexico City, Mexico: Association for Computational Linguistics.

\bibitem[{Team et~al.(2025)Team, Kamath, Ferret, Pathak, Vieillard, Merhej, Perrin, Matejovicova, Ram{\'e}, Rivi{\`e}re et~al.}]{team2025gemma}
Team, G.; Kamath, A.; Ferret, J.; Pathak, S.; Vieillard, N.; Merhej, R.; Perrin, S.; Matejovicova, T.; Ram{\'e}, A.; Rivi{\`e}re, M.; et~al. 2025.
\newblock Gemma 3 technical report.
\newblock \emph{arXiv preprint arXiv:2503.19786}.

\bibitem[{Van and Wu(2023)}]{vanwu2023vlm}
Van, M.-H.; and Wu, X. 2023.
\newblock Detecting and Correcting Hate Speech in Multimodal Memes with Large Visual Language Model.
\newblock In \emph{arXiv preprint arXiv:2311.06737}.

\bibitem[{Wang et~al.(2025)Wang, Ni, Zhang, Lu, Hu, He, Hu, Lin, Guo, Du et~al.}]{wang2025repomaster}
Wang, H.; Ni, Z.; Zhang, S.; Lu, S.; Hu, S.; He, Z.; Hu, C.; Lin, J.; Guo, Y.; Du, Y.; et~al. 2025.
\newblock RepoMaster: Autonomous Exploration and Understanding of GitHub Repositories for Complex Task Solving.
\newblock \emph{arXiv preprint arXiv:2505.21577}.

\bibitem[{Wang et~al.(2024)Wang, Zhang, Fei, Chen, and Qin}]{Wang2024S3Agent}
Wang, P.; Zhang, Y.; Fei, H.; Chen, Q.; and Qin, L. 2024.
\newblock S3 Agent: Unlocking the Power of VLLM for Zero-Shot Multi-Modal Sarcasm Detection.
\newblock \emph{ACM Transactions on Multimedia Computing, Communications, and Applications}.

\bibitem[{Wang et~al.(2023)Wang, Mao, Wu, Ge, Wei, and Ji}]{wang2023unleashing}
Wang, Z.; Mao, S.; Wu, W.; Ge, T.; Wei, F.; and Ji, H. 2023.
\newblock Unleashing the emergent cognitive synergy in large language models: A task-solving agent through multi-persona self-collaboration.
\newblock \emph{arXiv preprint arXiv:2307.05300}.

\bibitem[{Wu et~al.(2023)Wu, Bansal, Zhang, Wu, Li, and {\em et al.}}]{Wu2023AutoGen}
Wu, Q.; Bansal, G.; Zhang, J.; Wu, Y.; Li, B.; and {\em et al.} 2023.
\newblock AutoGen: Enabling Next‑Gen {LLM} Applications via Multi‑Agent Conversation.
\newblock \emph{arXiv preprint}, arXiv:2308.08155.

\bibitem[{Xuan et~al.(2024)Xuan, Yi, Yang, Wu, Fung, and Ji}]{xuan2024lemma}
Xuan, K.; Yi, L.; Yang, F.; Wu, R.; Fung, Y.~R.; and Ji, H. 2024.
\newblock LEMMA: LVLM-Enhanced Multimodal Misinformation Detection with External Knowledge Augmentation.
\newblock \emph{arXiv preprint arXiv:2402.11943}.

\bibitem[{Yamagishi(2024)}]{yamagishi2024zeroshot}
Yamagishi, Y. 2024.
\newblock Simpler Prompts, Better Results: Enhancing Zero-shot Detection with a Large Multimodal Model.
\newblock In \emph{Proceedings of CASE 2024}.

\bibitem[{Yoffe, Amayuelas, and Wang(2024)}]{yoffe2024debunc}
Yoffe, L.; Amayuelas, A.; and Wang, W.~Y. 2024.
\newblock DebUnc: Improving Large Language Model Agent Communication With Uncertainty Metrics.
\newblock \emph{arXiv preprint arXiv:2407.06426}.

\bibitem[{Zhang et~al.(2024)Zhang, Sun, Chen, Pfister, Zhang, and Arik}]{li2024chainofagents}
Zhang, Y.; Sun, R.; Chen, Y.; Pfister, T.; Zhang, R.; and Arik, S. 2024.
\newblock Chain of Agents: Large Language Models Collaborating on Long‑Context Tasks.
\newblock In \emph{Proceedings of the 38th Conference on Neural Information Processing Systems (NeurIPS 2024)}.

\bibitem[{Zhang et~al.(2025)Zhang, Zou, Wang, and Qin}]{Zhang2025CommanderGPT}
Zhang, Y.; Zou, C.; Wang, B.; and Qin, J. 2025.
\newblock Commander-GPT: Fully Unleashing the Sarcasm Detection Capability of Multi-Modal Large Language Models.
\newblock \emph{arXiv preprint arXiv:2503.18681}.

\end{thebibliography}
\end{document}